# Solving Influence Diagrams using HUGIN, Shafer-Shenoy and Lazy Propagation

**Anders L. Madsen**
Hugin Expert A/S
Niels Jernes Vej 10
Box 8201
DK-9220 Aalborg Ø, Denmark
Anders.L.Madsen@hugin.com

**Dennis Nilsson**
Department of Mathematical Sciences
Aalborg University
Fredrik Bajers Vej 7G
DK-9220 Aalborg Ø, Denmark
nilsson@math.auc.dk

## Abstract

In this paper we present three different architectures for the evaluation of influence diagrams: HUGIN, Shafer-Shenoy (S-S), and Lazy Propagation (LP). HUGIN and LP are two new architectures introduced in this paper. The computational complexity using the three architectures are compared on the same structure, the LImited Memory Influence Diagram (LIMID), where only the requisite information for the computation of optimal policies is depicted. Because the requisite information is explicitly represented in the diagram, the evaluation procedure can take advantage of it. Previously, it has been shown that significant savings in computational time can be obtained by performing the calculation on the LIMID rather than on the traditional influence diagram. In this paper we show how the obtained savings is considerably increased when the computations are performed according to the LP scheme.

## 1 Introduction

In the last decade, several architectures have been proposed for the evaluation of influence diagrams. The pioneering architecture was proposed by Olmsted (1983), and Shachter (1986). Their method works directly on the influence diagram by eliminating the nodes from the diagram in a reverse time ordering. Shenoy (1992) proposed an alternative formulation, *valuation based systems,* for the representation and evaluation of such decision problems. Later on, Jensen *et al.* (1994) described an algorithm that solves influence diagrams by the propagation of messages in so-called *strong junction trees.*

Recently, Lauritzen and Nilsson (1999) introduced the notion of LImited Memory Influence Diagrams (LIMIDs) to describe multistage decision problems, and presented a procedure, termed Single Policy Updating (SPU), for evaluating them. In contrast with traditional influence diagrams, LIMIDs allow for the possibility of violating the 'no-forgetting' assumption. Thus, in particular any influence diagram can be represented as a LIMID whereas the converse does not hold in general. In Nilsson and Lauritzen (2000), it is shown how SPU applied on influence diagrams, can yield significant savings in computational time when compared to traditional influence diagram algorithms. In the above paper, the computations performed during SPU were done by the passage of messages in a suitable junction tree using the S-S architecture.

In this paper we show how SPU for influence diagrams can be performed using two new architectures: The LP architecture which has some resemblance with the method described in Madsen (1999), and the Hugin architecture which has some resemblance with the method described in Jensen *et al.* (1994). A comparison of the computational efficiency of the three architectures is then presented.

## 2 LIMIDs

A LIMID is a directed acyclic graph consisting of three types of nodes: *Chance nodes* representing random variables, *decision nodes* representing decisions to be taken, and *value nodes* representing (local) utility functions. The three types of nodes are represented as circles, boxes, and diamonds, respectively. The set of chance nodes is denoted $\Gamma$, the set of decision nodes is denoted by $\Delta$, and the set of value nodes is denoted by $\Upsilon$.

The arcs in the LIMID have a different meaning depending on their destination. Arcs into chance nodes represent probabilistic dependence, and associated with chance node $r$ is a conditional probability function $p_r$ of the variable given its parents. Arcs into decision nodes are *informational*, and the parents of decision node $d$ are the variables whose values are known to the decision maker at the time the decision $d$ must be taken. Arcs into value nodes indicate functional dependence, and the parents of value node $u$ are the variables that the local utility function $U_u$ associ-ated with $u$ depends on.

In contrast with traditional influence diagrams, the informational arcs in LIMIDs are not restricted to obey the 'no-forgetting' assumption. This assumption states that an observation made prior to a given decision must be known to the decision maker on all subsequent decisions. Since no-forgetting is not assumed in LIMIDs, the LIMID evaluation algorithm (SPU) can take advantage of this flexibility, by removing informational arcs into decision nodes that are not neccessary for the computation of the optimal strategy. The removal of informational arcs are determined solely from the structure of the LIMID, and is performed prior to any numerical evaluation of the LIMID.

## 2.1 Strategies

We let $V = \Gamma \cup \Delta$. Elements in $V$ will be termed *variables* or *nodes* interchangeably. The variable $n \in V$ can take values in a finite set $\mathcal{X}_n$. For $W \subseteq V$, we let $\mathcal{X}_W = \times_{n \in W} \mathcal{X}_n$. Elements of $\mathcal{X}_W$ are denoted by lower case letters such as $x_W$, abbreviating $x_V$ to $x$. The set of parents of a node $n$ is denoted $\text{pa}(n)$. The *family* of $n$, denoted $\text{fa}(n)$, is defined by $\text{fa}(n) = \text{pa}(n) \cup \{n\}$.

A *policy* for decision node $d$ is a function $\delta_d$ that associates with each state $x_{\text{pa}(d)}$ a probability distribution $\delta_d(\cdot \mid x_{\text{pa}(d)})$ on $\mathcal{X}_d$. A *uniform policy* for $d$, denoted $\bar{\delta}_d$, is given by $\bar{\delta}_d = 1/|\mathcal{X}_d|$. A *strategy* $q$ is a collection of policies $q = \{\delta_d : d \in \Delta\}$, one for each decision. The strategy $q$ induces a joint distribution of all the variables in $V$ as

$$f_q = \prod_{r \in \Gamma} p_r \prod_{d \in \Delta} \delta_d. \quad (1)$$

The expected utility of a strategy $q$ is the expectation of the total utility $U = \sum_{u \in \Upsilon} U_u$ wrt. the joint distribution of $V$ induced by $q$: $\text{EU}(q) = \sum_x f_q(x) U(x)$. A *global maximum strategy*, or simply an *optimal strategy*, denoted $\hat{q}$, is a strategy that maximizes the expected utility, i.e. $\text{EU}(\hat{q}) \geq \text{EU}(q)$ for all strategies $q$. The individual policies in an optimal strategy are termed *optimal*.

## 2.2 Single Policy Updating

SPU is an iterative procedure for evaluating general LIMIDs. The procedure starts with an initial strategy and improves it by local updates until convergence has occurred, i.e. until every local change would result in an inferior strategy. Given a strategy $q = \{\delta_d : d \in \Delta\}$ and $d_0 \in \Delta$, we let $q_{-d_0} = q \setminus \{\delta_{d_0}\}$ be the partially specified strategy obtained by retracting the policy for $d_0$ from $q$.

SPU starts with an initial strategy and proceeds by modifying (updating) the policies in a random or systematically order. If the current strategy is $q$ and the policy for $d_i$ is to be updated, then the following steps are performed:

- **Retract:** Retract the current policy for $d_i$ from $q$ to obtain $q_{-d_i}$.

- **Optimize:** Compute a new policy for $d_i$ by:

$$\tilde{\delta}_{d_i} = \arg\max_{\delta_{d_i}} \text{EU}\left(q_{-d_i} \cup \{\delta_{d_i}\}\right).$$

- **Replace:** Redefine $q := q_{-d_i} \cup \{\tilde{\delta}_{d_i}\}$.

The policies are updated until they converge to a strategy in which no single policy modification can increase the expected utility.

## 2.3 Single Policy Updating for Influence Diagrams

Suppose we apply SPU on a traditional influence diagram with decision nodes $\Delta = \{d_1, \ldots, d_k\}$, where the index of the decisions indicate the order in which they are to be taken, i.e. $d_1$ is the initial decision, and $d_k$ is the last decision to be taken. In this case, we always compute an optimal strategy using SPU, if we

- start with the uniform policies on all the decisions;
- update the policies for the decisions using the order $d_k, \ldots, d_1$.

Furthermore, the optimal strategy is computed after exactly one update of each policy.

When the policy for $d_i$ is to be updated, the optimal policy for $d_i$ is found by computing (letting $d = d_i$)

$$\delta_d^* = \arg\max_{\delta_d} \sum_{V \setminus \text{fa}(d)} \left\{ \delta_d (\prod_{r \in \Gamma} p_r)(\prod_{j=i+1}^{k} \delta_{d_j}^*)(\sum_{u \in \Upsilon} U_u) \right\}, \quad (2)$$

where $\delta_{d_j}^*$ are optimal policies for $d_j, j = i+1, \ldots, k$.

Note that in the expression (2), the uniform policies for $d_1, \ldots, d_{i-1}$ are not included. This is because they have no effect on the maximizing policy $\delta_{d_i}^*$.

## 2.4 Construction of the junction tree

SPU is performed efficiently in a computational structure called *junction tree*. We abstain here from explaining all the details in the compilation procedure, but refer to Lauritzen and Nilsson (1999) for further reading. In brevity, the compilation of a LIMID into a junction tree consists of four steps, see Fig. 1:

**Reduction:** Here, all *non-requiste* informational arcs into the decision nodes are removed. A non-requiste arc from a node $n$ into decision node $d$ has the property that there exists an optimal policy for $d$ that does not depend on $n$. Denote the obtained LIMID $\mathcal{L}_{\min}$.

**Moralization:** In this step, pair of parents of any node in $\mathcal{L}_{\min}$ are 'married' by inserting a link between them. Then, the diagram is made undirected, and finally utility nodes are removed. Denote the obtained undirected graph by $\mathcal{L}^m$.

**Triangulation:** Here $\mathcal{L}^m$ is triangulated to obtain $\mathcal{L}^t$.

**Construction of the junction tree:** In this final step, a junction tree is constructed whose nodes correspond to the cliques of $\mathcal{L}^t$.

In the Triangulation step it is important to note that any triangulation order may be used. As a nice consequence, our junction tree is typically smaller than the strong junction tree as described in Jensen et al. (1994).

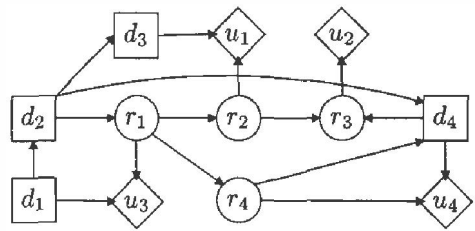

Figure 1: A reduced LIMID $\mathcal{L}$.

### 2.5 Partial collect propagation

Suppose we are given the junction tree representation $\mathcal{T}$ of a LIMID $\mathcal{L}$ with decision nodes $d_1, \ldots, d_k$. Initially, we update the policy for the last decision $d_k$ by passing messages from the leaves of $\mathcal{T}$ towards any one clique containing $\text{fa}(d_k)$. After this 'collect operation', a local optimization is performed to compute the updated policy for $d_k$. The local optimization processes differ slightly for the three architectures and are explained later. Next, the policy for decision $d_{k-1}$ is updated, and we could apply the same algorithm for this purpose. This procedure suggests that we have to perform $k$ collect operations in the course of the evaluation of the LIMID. The procedure usually involves a great deal of duplication. After the first collect operation towards any clique, say $R_k$, containing $\text{fa}(d_k)$, we must collect messages towards any one clique, say $R_{k-1}$, containing $\text{fa}(d_{k-1})$. However, because some of the messages have already been performed during the first collect operation, we need only pass messages from $R_k$ towards $R_{k-1}$. Thus, it can be seen that we need only perform one 'full' collect (towards $R_k$), and $k - 1$ 'partial' collect operations, see Fig. 2 for an illustration of the partial collect algorithm.

## 3 The Shafer-Shenoy architecture

In our local propagation scheme, the utilities and probabilities specified in the LIMID $\mathcal{L}$ are represented by entities called potentials:

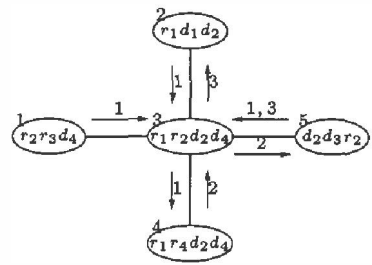

Figure 2: Message passing in the junction tree for $\mathcal{L}$.

**Definition 1 [Potential]**
A *potential* on $W \subseteq V$ is a pair $\pi_W = (p_W, u_W)$ where $p_W$ is a non-negative real function on $\mathcal{X}_W$, and $u_W$ is a real function on $\mathcal{X}_W$.

Thus, a potential consists of two parts: The first part is called the *probability part*, and the second part is termed the *utility part*. We call a potential $\pi_W$ *vacuous*, if $\pi_W = (1, 0)$. To represent and evaluate the decision problem in terms of potentials, we define basic operations of *combination* and *marginalization*:

**Definition 2 [Combination]**
The *combination* of two potentials $\pi_{W_1} = (p_{W_1}, u_{W_1})$ and $\pi_{W_2} = (p_{W_2}, u_{W_2})$ denotes the potential on $W_1 \cup W_2$ given by $\pi_{W_1} \otimes \pi_{W_2} = (p_{W_1} p_{W_2}, u_{W_1} + u_{W_2})$.

**Definition 3 [Marginalization]**
The *marginalization* of $\pi_W = (p_W, u_W)$ onto $W_1 \subseteq W$ is defined by

$$\pi_W^{\downarrow W_1} = \left( \sum_{W \setminus W_1} p_W \, , \, \frac{\sum_{W \setminus W_1} p_W u_W}{\sum_{W \setminus W_1} p_W} \right).$$

Here we have used the convention that $0/0 = 0$ which will be used throughout.

As shown in Lauritzen and Nilsson (1999), the operations of combination and marginalization satisfy the properties of S-S axioms (see Shenoy and Shafer (1990)). This establishes the correctness of the propagation algorithm presented in Theorem 1.

### 3.1 Initialization

To initialize the junction tree $\mathcal{T}$ one first associates a vacuous potential to each clique $C \in \mathcal{C}$. Then, for each chance node, $r$, $p_r$ is multiplied onto the probability part of any clique $C$ satisfying $C \supseteq \text{fa}(r)$. Similarly, for each decision node, $d$, the uniform policy $\bar{\delta}_d$ is multiplied into the probability part of any clique $C$ satisfying $C \supseteq \text{fa}(d)$. Finally, for each value node $u$, $U_u$ is added onto the utility part of any clique $C$ satisfying $C \supseteq \text{fa}(u)$. The compilation process of the LIMID into $\mathcal{T}$ guarantees the existence of such cliques. Since we start with uniform policies it is unnecessary to include policies in the initialization.

Let $\pi_C = (p_C, u_C)$ be the potential on $C$ after these operations have been performed. The *joint potential* $\pi_V$ is the combination of all the potentials and satisfies

$$\pi_V = (p_V, u_V) = \otimes_{C \in \mathcal{C}} \pi_C = \left( \prod_{r \in \Gamma} p_r, \sum_{u \in \Upsilon} U_u \right). \quad (3)$$

So, we may write the updating policy in (2) shortly as

$$\delta^*_{d_i} = \arg\max_{\delta_{d_i}} \sum_{V \setminus \mathrm{fa}(d_i)} \left\{ p_V u_V \, \delta_{d_i} \prod_{j=i+1}^{k} \delta^*_{d_j} \right\}. \quad (4)$$

### 3.2 Message passing

Suppose we wish to find the marginal $\pi_V^{\downarrow C}$ of some clique $C$. To achieve our purpose we pass messages in $\mathcal{T}$ via a pair of *mailboxes* placed on each edge in the junction tree. The mailboxes between two neighbours $A$ and $B$ can contain potentials on $A \cap B$.

Now, we direct all the edges in $\mathcal{T}$ towards the clique $C$. Then, each node pass a message towards its child whenever the sender has received messages from all its parents. When a message is sent from $A$ to $B$, we insert a message $\pi_{A \to B}$ in the mailbox given by

$$\pi_{A \to B} = \left( \pi_A \otimes \left( \otimes_{C \in \mathrm{ne}(A) \setminus \{B\}} \pi_{C \to A} \right) \right)^{\downarrow B},$$

where $\mathrm{ne}(A)$ is the set of neighbours of $A$.

**Theorem 1** *Suppose we start with a joint potential $\pi_V$ on a junction tree $\mathcal{T}$, and pass messages towards a 'root-clique' $R$ as described above. When $R$ has received a message from each of its neighbours, the combination of all messages with its own potential is equal to the $R$-marginal of the joint potential $\pi_V$:*

$$\pi_V^{\downarrow R} = (\otimes_{C \in \mathcal{C}} \pi_C)^{\downarrow R} = \pi_R \otimes \left( \otimes_{C \in \mathrm{ne}(R)} \pi_{C \to R} \right),$$

*where $\mathcal{C}$ is the set of cliques in $\mathcal{T}$.*

### 3.3 Local optimization

This section is concerned with showing how SPU is performed by message passing in the junction tree $\mathcal{T}$.

Letting the *contraction* $\mathrm{cont}(\pi_W)$ of a potential $\pi_W = (p_W, u_W)$ be the real valued function on $\mathcal{X}_W$ given as $\mathrm{cont}(\pi_W) = p_W u_W$ it is easily shown that for $W_1 \subseteq W$ we have

$$\mathrm{cont}(\pi_W^{\downarrow W_1}) = \sum_{W \setminus W_1} \mathrm{cont}(\pi_W). \quad (5)$$

Suppose we want to update the policy for $d_i$ after having updated the policies for $d_{i+1}, \ldots, d_k$ and obtained $\delta^*_{d_{i+1}}, \ldots, \delta^*_{d_k}$, and assume the joint potential on $\mathcal{T}$ is

$$\pi^*_V = \left( p_V \prod_{j=i+1}^{k} \delta^*_{d_j}, u_V \right).$$

Then, according to (4), (5), and Theorem 1, the updated policy $\delta^*_{d_i}$ for $d_i$ can be found by carrying out the following steps (abbreviating $d_i$ into $d$):

1. **Collect:** Collect to any clique $R$ containing $\mathrm{fa}(d)$ to obtain $\pi^*_R = (\pi^*_V)^{\downarrow R}$.

2. **Marginalize:** Compute $\pi^*_{\mathrm{fa}(d)} = (\pi^*_R)^{\downarrow \mathrm{fa}(d)}$.

3. **Contract:** Compute the contraction $c_{\mathrm{fa}(d)}$ of $\pi^*_{\mathrm{fa}(d)}$.

4. **Optimize:** Define $\delta^*_d(x_{\mathrm{pa}(d)})$ for all $x_{\mathrm{pa}(d)}$ as the distribution degenerate at a point $x^*_d$ satisfying

$$x^*_d = \arg\max_{x_d} c_{\mathrm{fa}(d)}(x_d, x_{\mathrm{pa}(d)}).$$

When the above steps have been performed, the policy $\delta^*_{d_i}$ is multiplied onto the probability part of $R$ such that the joint potential on $\mathcal{T}$ becomes

$$\pi^*_V = \left( p_V \prod_{j=i}^{k} \delta^*_{d_j}, u_V \right).$$

Now, we can in a similar manner update the policies for $d_{i-1}, \ldots, d_1$. When all policies have been updated in this way, the obtained strategy $(\delta^*_{d_1}, \ldots, \delta^*_{d_k})$ is optimal.

## 4 Lazy Propagation architecture

The LP architecture is based on maintaining decompositions of potentials. Therefore, a generalized notion of potentials is introduced.

**Definition 4 [Potential]**
A *potential* on $W \subseteq V$ is a pair $\pi = (\Phi, \Psi)$ where $\Phi$ is a set of non-negative real functions on subsets of $\mathcal{X}_W$, and $\Psi$ is a set of real functions on subsets of $\mathcal{X}_W$.

Thus, the probability part of a potential is a set of probability functions and policies whereas the utility part is a set of local utility functions. We call a potential $\pi_W$ *vacuous*, if $\pi_W = (\emptyset, \emptyset)$. We define new basic operations of *combination* and *marginalization*:

**Definition 5 [Combination]**
The *combination* of two potentials $\pi_{W_1} = (\Phi_1, \Psi_1)$ and $\pi_{W_2} = (\Phi_2, \Psi_2)$ denotes the potential on $W_1 \cup W_2$ given by $\pi_{W_1} \otimes \pi_{W_2} = (\Phi_1 \cup \Phi_2, \Psi_1 \cup \Psi_2)$.

**Definition 6 [Marginalization]**
The *marginalization* of $\pi_W = (\Phi, \Psi)$ onto $W \setminus W_1$ is defined by

$$\pi_W^{\downarrow W_1} = \left( \Phi \setminus \Phi_{W_1} \cup \{\phi_{W_1}^*\}, \Psi \setminus \Psi_{W_1} \cup \left\{\frac{\psi_{W_1}^*}{\phi_{W_1}^*}\right\} \right),$$

where

$$\phi_{W_1}^* = \sum_{W \setminus W_1} \prod_{\phi \in \Phi_{W_1}} \phi,$$
$$\psi_{W_1}^* = \sum_{W \setminus W_1} \left( \prod_{\phi \in \Phi_{W_1}} \phi \sum_{\psi \in \Psi_{W_1}} \psi \right),$$

$\Phi_{W_1} = \{\phi \in \Phi \mid W_1 \cap dom(\phi) \neq \emptyset\}$, and $\Psi_{W_1} = \{\psi \in \Psi \mid W_1 \cap dom(\psi) \neq \emptyset\}$.

The above operations of combination and marginalization satisfy the properties of the S-S axioms, see Shafer and Shenoy (1990).

**Lemma 1 (Commutativity and Associativity)**
Let $\pi_{W_1}$, $\pi_{W_2}$, and $\pi_{W_3}$ be potentials. Then

$$\pi_{W_1} \otimes \pi_{W_2} = \pi_{W_2} \otimes \pi_{W_1} \text{ and}$$
$$\pi_{W_1} \otimes (\pi_{W_2} \otimes \pi_{W_3}) = (\pi_{W_1} \otimes \pi_{W_2}) \otimes \pi_{W_3}.$$

Lemma 1 allows us to use the notation $\pi_{W_1} \otimes \pi_{W_2} \otimes \pi_{W_3}$.

**Lemma 2 (Consonance)**
Let $\pi_W$ be a potential on $W$, and let $W \supseteq W_1 \supseteq W_2$. Then $(\pi_W^{\downarrow W_1})^{\downarrow W_2} = \pi_W^{\downarrow W_2}$.

**Lemma 3 (Distributivity)**
Let $\pi_{W_1}$ and $\pi_{W_2}$ be potentials on $W_1$ and $W_2$, respectively. Then $(\pi_{W_1} \otimes \pi_{W_2})^{\downarrow W_1} = \pi_{W_1} \otimes \pi_{W_2}^{\downarrow W_1}$.

Lemmas 1-3 establish the correctness of the propagation algorithm presented in Theorem 2.

### 4.1 Initialization

The initialization of the junction tree proceeds as in the S-S architecture with the exception that the probability, policy, and utility functions associated with a clique are not combined. Thus, after initialization each clique $C$ holds a potential $\pi_C = (\Phi, \Psi)$. Since we start with uniform policies it is unnecessary to include policies in the initialization. Notice, that $\bigcup_{\phi \in \Phi} dom(\phi), \bigcup_{\psi \in \Psi} dom(\psi) \subseteq C$.

Let $\pi_C = (\Phi, \Psi)$ be the potential on clique $C$ after initialization. The joint potential $\pi_V = (\Phi_V, \Psi_V) = \otimes_{C \in \mathcal{C}} \pi_C$ on $\mathcal{T}$ is the combination of all potentials and satisfies $\pi_V = (\Phi_V, \Psi_V) = (\{p_r : r \in \Gamma\}, \{U_u : u \in \Upsilon\})$.

The updating policy in (2) may be written as

$$\delta_{d_i}^* = \arg\max_{\delta_{d_i}} \sum_{V \setminus \text{fa}(d_i)} \left\{ \prod_{\phi \in \Phi_V} \phi \left( \sum_{\psi \in \Psi_V} \psi \right) \prod_{j=i}^{k} \delta_{d_j}^* \right\}. \quad (6)$$

### 4.2 Message passing

Messages are passed between the cliques of $\mathcal{T}$ via mailboxes as in the S-S architecture. Let $\{\pi_C : C \in \mathcal{C}\}$ be the collection of potentials on $\mathcal{T}$. The passage of a message $\pi_{A \to B}$ from clique $A$ to clique $B$ is performed by absorption. Absorption from clique $A$ to clique $B$ involves eliminating the variables $A \setminus B$ from the potentials associated with $A$ and its neighbours except $B$. The structure of the message $\pi_{A \to B}$ is given by

$$\pi_{A \to B} = \left( \pi_A \otimes \left( \otimes_{C \in ne(A) \setminus \{B\}} \pi_{C \to A} \right) \right)^{\downarrow B},$$

where $ne(A)$ are the neighbours of $A$ in $\mathcal{T}$ and $\pi_{C \to A}$ is the message passed from $C$ to $A$.

**Theorem 2** *Suppose we start with a joint potential $\pi_V$ on a junction tree $\mathcal{T}$, and pass messages towards a 'root clique' $R$ as described above. When $R$ has received a message from each of its neighbours, the combination of all messages with its own potential is equal to a decomposition of the $R$-marginal of $\pi_V$:*

$$\pi_V^{\downarrow R} = (\otimes_{C \in \mathcal{C}} \pi_C)^{\downarrow R} = \pi_R \otimes (\otimes_{C \in ne(R)} \pi_{C \to R}),$$

*where $\mathcal{C}$ is the set of cliques in $\mathcal{T}$.*

### 4.3 Local optimization

This section is concerned with showing how SPU is performed by message passing in the junction tree using LP. SPU in the LP architecture proceeds as SPU in the S-S architecture. The operations are, however, different. The contraction $\text{cont}(\pi_W)$ of a potential $\pi_W = (\Phi, \Psi)$ is the real function on $\mathcal{X}_W$ given as

$$\text{cont}(\pi_W) = \prod_{\phi \in \Phi} \phi \sum_{\psi \in \Psi} \psi.$$

As in S-S, it is easily shown that for $W_1 \subseteq W$ we have

$$\text{cont}(\pi_W^{\downarrow W_1}) = \sum_{W \setminus W_1} \text{cont}(\pi_W). \quad (7)$$

Let $\pi_C = (\Phi, \Psi)$ be the clique potential for clique $C$. The domain of the contraction of $\pi_C$ is

$$dom(\text{cont}(\pi_C)) = \bigcup_{\phi \in \Phi} dom(\phi) \cup \bigcup_{\psi \in \Psi} dom(\psi)$$

and has the property $dom(\text{cont}(\pi_C)) \subseteq C$.

Assume we want to update the policy for $d_i$ after having updated the policies for $d_{i+1}, \ldots, d_k$. Further, assume the joint potential on $\mathcal{T}$ is

$$\pi_V^* = \left( \Phi_V \cup \{\delta_{d_{i+1}}^*, \ldots, \delta_{d_k}^*\}, \Psi_V \right).$$

Then, according to (6), (7), and Theorem 2, the updated policy $\delta_{d_i}^*$ for $d_i$ can be found by carrying out the steps of Section 3.3 using the operations of the LP architecture. When these steps have been performed, the policy $\delta_{d_i}^*$ is assigned to the probability part of $R$ such that the joint potential on $\mathcal{T}$ becomes

$$\pi_V^* = \left( \Phi_V \cup \{\delta_{d_i}^*, \ldots, \delta_{d_k}^*\}, \Psi_V \right).$$

The policies for $d_{i-1}, \ldots, d_1$ are updated in a similar manner. Once all policies have been updated, the obtained strategy $(\delta_{d_1}^*, \ldots, \delta_{d_k}^*)$ is a global maximum strategy.

### 4.4 Local computation

When computing the message $\pi_{A \to B}$, the marginalization of $A \setminus B$ can be performed efficiently by local computation, if variables are eliminated one at a time. By eliminating variables one at a time, barren variables can be exploited.

**Definition 7** A variable $n$ is a *barren variable* wrt. a potential $\pi_V = (\Phi, \Psi)$ and a set $W$, if neither $n \in W$ nor $n \in dom(\psi)$ for any $\psi \in \Psi$, and $n$ only has barren descendants in the domain graph of $\pi_V$. A *probabilistic barren variable* wrt. $\pi_V$ and $W$ is a variable which is barren when only $\Phi$ is considered.

The marginalization of a barren variable $n$ from a potential $\pi_W$ when computing a message produces a vacuous conditional probability function $\phi_n^*$ which is not included in the probability part of $\pi_W^{\downarrow W \setminus \{n\}}$. Whether or not $n$ is a barren variable can be detected efficiently by a structural analysis on the domain graph of $\pi$ before the marginalization operations are performed. A barren variable can be eliminated without computation since it does not contribute with any information to $\pi_{A \to B}$. Similarly, a probabilistic barren variable does not contribute with any information to the probability part of $\pi_{A \to B}$.

## 5 The HUGIN architecture

The final architecture to be considered is HUGIN. It differs from both the S-S architecture and the LP architecture in the representation of the joint potential and also in the messages passed. With each pair of neighbours $A$ and $B$ in $\mathcal{T}$ we associate the separator $S = A \cap B$. The set of separators $\mathcal{S}$ play an explicit role in the HUGIN architecture as they themselves hold potentials $\pi_S$, $S \in \mathcal{S}$.

In addition to the combination and marginalization as defined in Section 3, we define a third operation on potentials:

**Definition 8** [Division]
The *division* between two potentials $\pi_A = (p_A, u_A)$ and $\pi_B = (p_B, u_B)$ is defined as

$$\pi_A \ominus \pi_B = \left( \frac{p_A}{p_B}, u_A - u_B \right).$$

### 5.1 Initialization

The initialization of the junction tree proceeds as in the S-S architecture. In addition, the separator $S \in \mathcal{S}$ is initialized with a vacuous potential. When the tree is initialized, the joint potential $\pi_V$ is given by

$$\pi_V = (\otimes_{C \in \mathcal{C}} \pi_C) \ominus (\otimes_{S \in \mathcal{S}} \pi_S) = \left( \prod_{r \in \Gamma} p_r, \sum_{u \in \Upsilon} U_u \right).$$

### 5.2 Message passing

The messages to be sent differ from the messages in the previous architectures by exploiting the separator potentials directly. Suppose that prior to a message is passed from $A$ to $B$ across the separator $S$, the potentials are $\pi_A$, $\pi_B$, and $\pi_S$ respectively. After the message is passed, the potentials change as $\pi_A^* = \pi_A$, $\pi_S^* = \pi_A^{\downarrow S}$, and $\pi_B^* = \pi_B \otimes (\pi_S^* \ominus \pi_S)$. The reader may easily verify that

$(\pi_A \otimes \pi_B) \ominus \pi_S = (\pi_A^* \otimes \pi_B^*) \ominus \pi_S^*$, from which it can be seen that the joint potential is unchanged under message passing.

The standard proofs in Dawid (1992) and Nilsson (2001) can be used directly to show that after collecting messages towards a node $R$ in the junction tree, $R$ will hold a potential from which the contraction of $\pi_V^{\downarrow R}$ can be computed:

**Theorem 3** *Suppose we start with a joint potential $\pi_V$ on a junction tree, and collect messages towards an arbitrary 'root-clique' $R$ as described above. Then, the resulting potential $\pi_R^* = (p_R^*, u_R^*)$ on $R$ satisfies*

$$p_R^* u_R^* = cont\left( \pi_V^{\downarrow R} \right).$$

Notice the difference between Theorem 3 and Theorems 1 and 2 which is due to the way, the theorems are proved.

### 5.3 Local optimization

The contraction cont($\pi_W$) of a potential $\pi_W$ is defined in the same way as in the S-S architecture. Further, when decision $d_i$ is to be updated in SPU, the steps to be carried out in the HUGIN architecture differ slightly from the previous architectures. According to Theorem 3, and (5), the updated policy $\delta^*_{d_i}$ for $d_i$ can be found by carrying out the following steps (abbreviating $d_i$ into $d$):

1. **Collect:** Collect to any clique $R$ containing fa($d$) to obtain $\pi^*_R = (p^*_R, u^*_R)$.
2. **Contract:** Compute the contraction $c_R$ of $\pi^*_R$.
3. **Marginalize:** Compute $c_{\text{fa}(d)} = \sum_{R \setminus \text{fa}(d)} c_R$.
4. **Optimize:** Define $\delta^*_d(x_{\text{pa}(d)})$ for all $x_{\text{pa}(d)}$ as the distribution degenerate at a point $x^*_d$ satisfying

$$x^*_d = \arg\max_{x_d} c_{\text{fa}(d)}(x_d, x_{\text{pa}(d)}).$$

## 6 Comparison

The message passing proceeds in the same way in each of the three architectures and it is for the LIMID $\mathcal{L}$ (Fig. 1) indicated in Fig. 2. The number of operations performed in each architecture during message passing is, however, different. The difference in the number of operations performed when solving $\mathcal{L}$ is shown in Tab. 1 assuming each variable to be binary.

| Algorithm | Sums | Mults | Divs | Subs | Total |
|---|---|---|---|---|---|
| S-S | 390 | 346 | 40 | 0 | 776 |
| HUGIN | 254 | 256 | 60 | 20 | 590 |
| LP | 170 | 180 | 16 | 0 | 280 |

Table 1: The number of operations performed for each algorithm when solving the LIMID $\mathcal{L}$ shown in Fig. 1.

Tab. 1 shows that the number of operations performed in the S-S architecture (776) is larger than the number of operations performed in the HUGIN architecture (590) which again is larger than the number of operations performed in the LP architecture (280) when solving $\mathcal{L}$. The numbers in Tab. 1 do not include the operations required for initialization. The initialization of the junction tree structure proceeds in the same manner for both the S-S and the HUGIN architectures. A straightforward implementation of the initialization would require 88 additional operations (48 multiplications and 40 additions). This excludes the combination of vacuous policies and clique probability potentials. In the LP architecture the junction tree is initialized when the probability potentials and utility functions have been assigned to cliques of the junction tree. This requires no additional operations and enables on-line exploitation of barren variables and independence relations.

### 6.1 Message passing

The fundamental differences between the LP, HUGIN, and S-S architectures can also be illustrated using $\mathcal{L}$. Consider the passing of the message $\pi_{32}$ from clique 3 to clique 2. Before this message is passed clique 3 has received messages $\pi_{13}, \pi_{43}, \pi_{53}$ from its other neighbours.

The message $\pi_{32}$ is in the S-S architecture computed using the following equation

$$\begin{aligned}\pi_{32} &= (\pi_3 \otimes \pi_{13} \otimes \pi_{43} \otimes \pi_{53})^{\downarrow r_1 d_2} \\ &= \left( \sum_{r_2}\sum_{d_4}(\phi_3\phi_{13}\phi_{43}\phi_{53}), \right. \\ & \left. \frac{\sum_{r_2}\sum_{d_4}\phi_3\phi_{13}\phi_{43}\phi_{53}(\psi_3 + \psi_{13} + \psi_{43} + \psi_{53})}{\sum_{r_2}\sum_{d_4}\phi_3\phi_{13}\phi_{43}\phi_{53}} \right).\end{aligned}$$ (8)

This equation requires a total of 164 operations. Described in words, to compute the message $\pi_{32}$, the combination of the clique potential $\pi_3$ of clique 3 and the incoming messages from neighbours of 3 except 2 is marginalized to $2 \cap 3 = \{r_1, d_2\}$. The potential $\pi_{32}$ is stored in the appropriate mailbox between 2 and 3.

The message $\pi_{32}$ is in the HUGIN architecture computed using the following equations

$$\begin{aligned}\pi_{32} &= (\pi_3)^{\downarrow r_1 d_2} \\ &= \left( \sum_{r_2}\sum_{d_4}\phi_3, \frac{\sum_{r_2}\sum_{d_4}\phi_3\psi_3}{\sum_{r_2}\sum_{d_4}\phi_3} \right).\end{aligned}$$ (9)

$$\begin{aligned}\pi^*_2 &= \pi_2 \otimes \pi_{32} \ominus \pi_{23} \\ &= \left( \phi_2\frac{\phi_{32}}{\phi_{23}}, \psi_2 + (\psi_{32} - \psi_{23}) \right).\end{aligned}$$ (10)

These equations require a total of 76 operations. Described in words, the message $\pi_{32}$ is computed by marginalization of $\pi_3$ to $2 \cap 3 = \{r_1, d_2\}$. Before updating the clique potential of $\pi_2$, the message $\pi_{32}$ is divided by the previous message $\pi_{23}$ passed in the opposite direction. The updated separator potential $\pi_{32}$ is stored in the mailbox between 2 and 3 replacing the previous separator potential $\pi_{23}$.

The message $\pi_{32}$ is in the LP architecture computed using equation 11 which requires a total of 60 operations. In words, to compute the message $\pi_{32}$, the combination of the clique potential $\pi_3$ of clique 3 and the incoming messages from neighbours of 3 except 2 is marginalized to $2 \cap 3 = \{r_1, d_2\}$. Thus, $r_2$ and $d_4$ have to be eliminated from the combination of potentials $\pi_3, \pi_{13}, \pi_{43}$, and

$$\begin{aligned}
\pi_{32} &= (\pi_3 \otimes \pi_{13} \otimes \pi_{43} \otimes \pi_{53})^{\downarrow r_1 d_2} \\
&= \left(\pi_3 \otimes \pi_{53} \oplus (\pi_{13} \otimes \pi_{43})^{\downarrow r_1 r_2 d_2}\right)^{\downarrow r_1 d_2} \\
&= \left(\{\}, \left\{\sum_{r_2} p(r_2|r_1)\Big(\psi(d_2,r_2) + \sum_{d_4} \phi(d_4|r_2,d_2)(\psi(r_2,d_4) + \psi(d_4,r_1,d_2))\Big)\right\}\right). \quad (11)
\end{aligned}$$

$\pi_{53}$. Since potentials are represented as decompositions it is possible to exploit independence relations between variables when computing messages. The domain graph $G$ induced by $\pi_3$, $\pi_{13}$, $\pi_{43}$, and $\pi_{53}$ is shown in Fig. 3. From $G$ it is readily seen that $r_2$ and $d_4$ are both probabilistic barren variables. This can be detected and exploited by the on-line triangulation algorithm. In the example, the order of elimination determined on-line when computing $\pi_{32}$ is $\sigma = <d_4, r_2>$. Since $d_4$ is only included in $\pi_{13}$ and $\pi_{43}$, $d_4$ is eliminated from their combination and subsequently $r_2$ is eliminated. The potential $\pi_{32}$ is stored in the appropriate mailbox between 2 and 3.

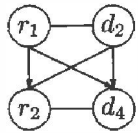

Figure 3: The domain graph of $\pi_3$, $\pi_{13}$, $\pi_{43}$, and $\pi_{53}$.

The number of operations to perform when computing the message $\pi_{32}$ is for each architecture shown in Tab. 2.

| Algorithm | Sums | Mults | Divs | Subs | Total |
|---|---|---|---|---|---|
| S-S | 88 | 80 | 4 | 0 | 164 |
| HUGIN | 32 | 32 | 8 | 4 | 76 |
| LP | 36 | 24 | 0 | 0 | 60 |

Table 2: The number of operations performed for each algorithm when sending the message $\pi_{32}$.

### 6.2 Architectural differences

The S-S architecture is a general architecture for inference in graphical models. It is based on the operations of marginalization and combination. Once the clique potentials are initialized, they are not changed during message passing.

The HUGIN architecture is a less general architecture than the S-S architecture since it requires a division operation. During message passing the clique potentials are updated by incoming messages. The division operation is introduced to avoid passing the information received from one clique back to the same clique. The update of clique potentials implies that in their basic form the HUGIN architecture is more efficient than the S-S architecture.

The LP architecture is based on operations of combination and marginalization. The architecture is similar to the S-S architecture in the sense that once initialized the clique potentials are not updated. The representation of the clique potentials is, however, different for the two architectures. Potentials are represented as decompositions. The elimination of a variable from a subset of potentials corresponds to on-line binarization of the secondary computational structure. This is a feature which is unique to the LP architecture. In the S-S and the HUGIN architectures the on-line order of elimination of variables does not have an impact on the computational requirements of computing a message. The domain graph induced by a potential is the complete graph and directions have been dropped. Therefore, the order of elimination is unimportant. In the LP architecture the on-line order of elimination is of high importance. Potentials are represented as decompositions and the direction of edges is maintained. This supports on-line exploitation of independence relations and directions of the LIMID from which the junction tree is constructed.

The LP architecture maintains decompositions of the probability and utility parts of potentials and postpone operations such that many unnecessary operations are avoided. Operations are performed during marginalization of variables and local optimization. The LP architecture may require repetition of already performed computations both when computing messages and eliminating variables. For instance, in the example the combination $\pi_{23} \otimes \pi_{13}$ is performed twice. It is performed when $\pi_{34}$ is computed and subsequently when $\pi_{35}$ is computed. This could be solved by binarization of the computational structure, but this increases the storage requirements. Binarization of the computational structure does not eliminate all repetitions of computations since variables are eliminated by local computation within each clique when computing messages. Thus, in order to eliminate all repetitions of computations some kind of nested caching structure like nested junction trees (Kjærulff (1998)), nested binary join trees (Shenoy (1997)), or a hash table of potentials is required. Binarization and nested caching structures reduce the degrees of freedom available when performing on-line triangulation. Basically, there is a time/space tradeoff with respect to the choice and granularity of the computational structure.

### 6.3 Evaluating general LIMIDs

The secondary computational structures of the S-S, HUGIN, and LP architectures are similar, but there are some important differences. The S-S and LP architectures use two mailboxes between each pair of neighbouring cliques to hold the most recent messages passed between the cliques. In HUGIN there is one mailbox between each pair of neighbouring cliques. This mailbox contains the most recent message passed between the two neighbouring cliques. In contrast with S-S and LP architectures, the HUGIN architecture multiplies the updated policy onto the clique potential. As a consequence the HUGIN architecture is not applicable for evaluating general LIMIDs. When evaluating general LIMIDs, we may have to update each policy more than once. In this case, we initially retract the policy to be updated from the joint potential before the updated policy is computed. This is easily done in LP, and can be done in S-S if we always store the policies separately in the cliques. However, as indicated above, the HUGIN architecture cannot retract policies, because messages are multiplied onto clique potentials during message passing.

## 7 Conclusion

The example LIMID $\mathcal{L}$ is not particularly complicated or has certain features. In fact, it is quite simple. Despite the simplicity of the example, there is a rather large difference in the number of operations required by each architecture to solve the LIMID. As can be seen from Tab. 1, LP offers an efficiency increase over HUGIN which offers an efficiency increase over S-S. We have compared the basic forms of the three architectures without applying any of the various speed-up techniques available (e.g. binary computational structures). This is done since each of the various speed-up techniques known by the authors applies equally well to each of the architectures.

In this paper we have introduced three different computational architectures for solving LIMID representations of influence diagrams. The HUGIN and the LP architectures are new whereas the S-S architecture was described in its original form in Nilsson and Lauritzen (2000). The differences between the architectures have been described in some detail. The LP architecture offers a way to reduce both the number of operations performed and the amount of space required during message passing. The LP architecture dissolves the difference between the HUGIN and S-S architectures. Finally, see Lepar and Shenoy (1998) for a similar comparison of HUGIN and S-S propagation, and Madsen and Jensen (1999) for a comparison of LP, and HUGIN and S-S architectures for computing marginals of probability distributions.

### Acknowledgement

Nilsson was supported by DINA (Danish Informatics Network in Agricultural Sciences), funded by the Danish Research Councils through their PIFT programme.